\documentclass{article}

\usepackage[final]{neurips_2024}

\usepackage{graphicx}
\usepackage[utf8]{inputenc} %
\usepackage[T1]{fontenc}    %
\usepackage{hyperref}       %
\usepackage{url}            %
\usepackage{booktabs}       %
\usepackage{amsfonts}       %
\usepackage{nicefrac}       %
\usepackage{microtype}      %
\usepackage{xcolor}         %
\usepackage{enumitem}
\usepackage{multirow}
\usepackage{tabularray}

\usepackage{xcolor}         %
\usepackage{hyperref}       %

\definecolor{uclablue}{rgb}{0.15, 0.45, 0.68} %
\hypersetup{
    breaklinks=true,
    colorlinks=true,
    linkcolor=uclablue,  
    citecolor=uclablue,  
    linktoc=page        
}

\title{
    \begin{tabular}{@{}lc@{}} %
        \setlength{\tabcolsep}{-100.0pt} %
        \multirow{2}{*}{\raisebox{-0.1\height}{\hspace{-10.0pt} \includegraphics[width=1.3cm]{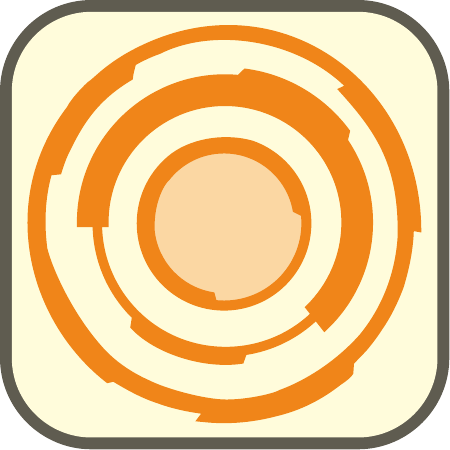}}} & 
        OS Agents: A Survey on MLLM-based Agents\\
        & for General Computing Devices Use\\
    \end{tabular}
}

\author{%
  \textbf{Xueyu Hu\textsuperscript{1, \dag},} \textbf{Tao Xiong\textsuperscript{1, \ddag},} \textbf{Biao Yi\textsuperscript{1, \ddag},} \textbf{Zishu Wei\textsuperscript{1, \ddag}}\\ \textbf{Ruixuan Xiao\textsuperscript{1}, Yurun Chen\textsuperscript{1}, Jiasheng Ye\textsuperscript{2}, Meiling Tao\textsuperscript{3}, Xiangxin Zhou\textsuperscript{4, 5}, Ziyu Zhao\textsuperscript{1},}\\ \textbf{Yuhuai Li\textsuperscript{1}, Shengze Xu\textsuperscript{6}, Shenzhi Wang\textsuperscript{7}, Xinchen Xu\textsuperscript{1}, Shuofei Qiao\textsuperscript{1}, Zhaokai Wang\textsuperscript{8}}\\ \textbf{Kun Kuang\textsuperscript{1}, Tieyong Zeng\textsuperscript{6}, Liang Wang\textsuperscript{4, 5}, Jiwei Li\textsuperscript{1}, Yuchen Eleanor Jiang\textsuperscript{3},}\\ \textbf{Wangchunshu Zhou\textsuperscript{3}, Guoyin Wang\textsuperscript{9}, Keting Yin\textsuperscript{1}, Zhou Zhao\textsuperscript{1},}\\ \textbf{Hongxia Yang\textsuperscript{10}, Fan Wu\textsuperscript{8}, Shengyu Zhang\textsuperscript{1, *}, Fei Wu\textsuperscript{1}}\\ \\ 
  \textsuperscript{1}Zhejiang University \textsuperscript{2}Fudan University \textsuperscript{3}OPPO AI Center\\ \textsuperscript{4}University of Chinese Academy of Sciences\\ \textsuperscript{5}Institute of Automation, Chinese Academy of Sciences\\ \textsuperscript{6}The Chinese University of Hong Kong \textsuperscript{7}Tsinghua University \textsuperscript{8}Shanghai Jiao Tong University \\ \textsuperscript{9}01.AI \textsuperscript{10}The Hong Kong Polytechnic University \\ \\ \{huxueyu, sy\_zhang\}@zju.edu.cn \\
  {\raisebox{-.2\height}{\includegraphics[width=0.6cm]{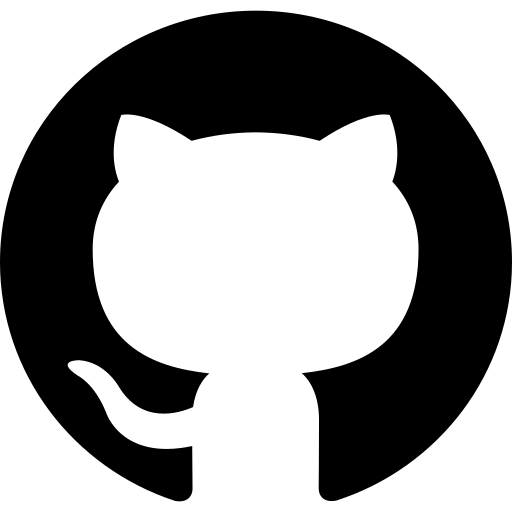}}}
\textbf{\url{https://os-agent-survey.github.io/}}\\
  {\raisebox{-.4\height}{\includegraphics[width=0.8cm]{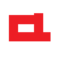}}}
\textbf{\url{https://aclanthology.org/2025.acl-long.369/}}  
}

\begin{document}
\maketitle

\renewcommand{\thefootnote}{}
\footnotetext{$^{\dag}$Projsssect Lead, $^{\ddag}$Core Contributor, $^{*}$Corresponding Author } 
\renewcommand{\thefootnote}{\arabic{footnote}}

\begin{abstract}
The dream to create AI assistants as capable and versatile as the fictional \textit{J.A.R.V.I.S} from \textit{Iron Man} has long captivated imaginations. With the evolution of (multi\-modal) large language models ((M)LLMs), this dream is closer to reality, as (M)LLM-based Agents using computing devices (e.g., computers and mobile phones) by operating within the environments and interfaces (e.g., Graphical User Interface (GUI)) provided by operating systems (OS) to automate tasks have significantly advanced. This paper presents a comprehensive survey of these advanced agents, designated as \textbf{OS Agents}. We begin by elucidating the fundamentals of OS Agents, exploring their key components including the environment, observation space, and action space, and outlining essential capabilities such as understanding, planning, and grounding. We then examine methodologies for constructing OS Agents, focusing on domain-specific foundation models and agent frameworks. A detailed review of evaluation protocols and benchmarks highlights how OS Agents are assessed across diverse tasks. Finally, we discuss current challenges and identify promising directions for future research, including safety and privacy, personalization and self-evolution. This survey aims to consolidate the state of OS Agents research, providing insights to guide both academic inquiry and industrial development. An open-source GitHub repository is maintained as a dynamic resource to foster further innovation in this field. We present a 9-page version of our work, accepted by ACL 2025, to provide a concise overview to the domain.
\end{abstract}

\newpage
{
  \tableofcontents
}
\clearpage

\begin{figure*}[!t]
    \centering
    \includegraphics[width=1\linewidth]{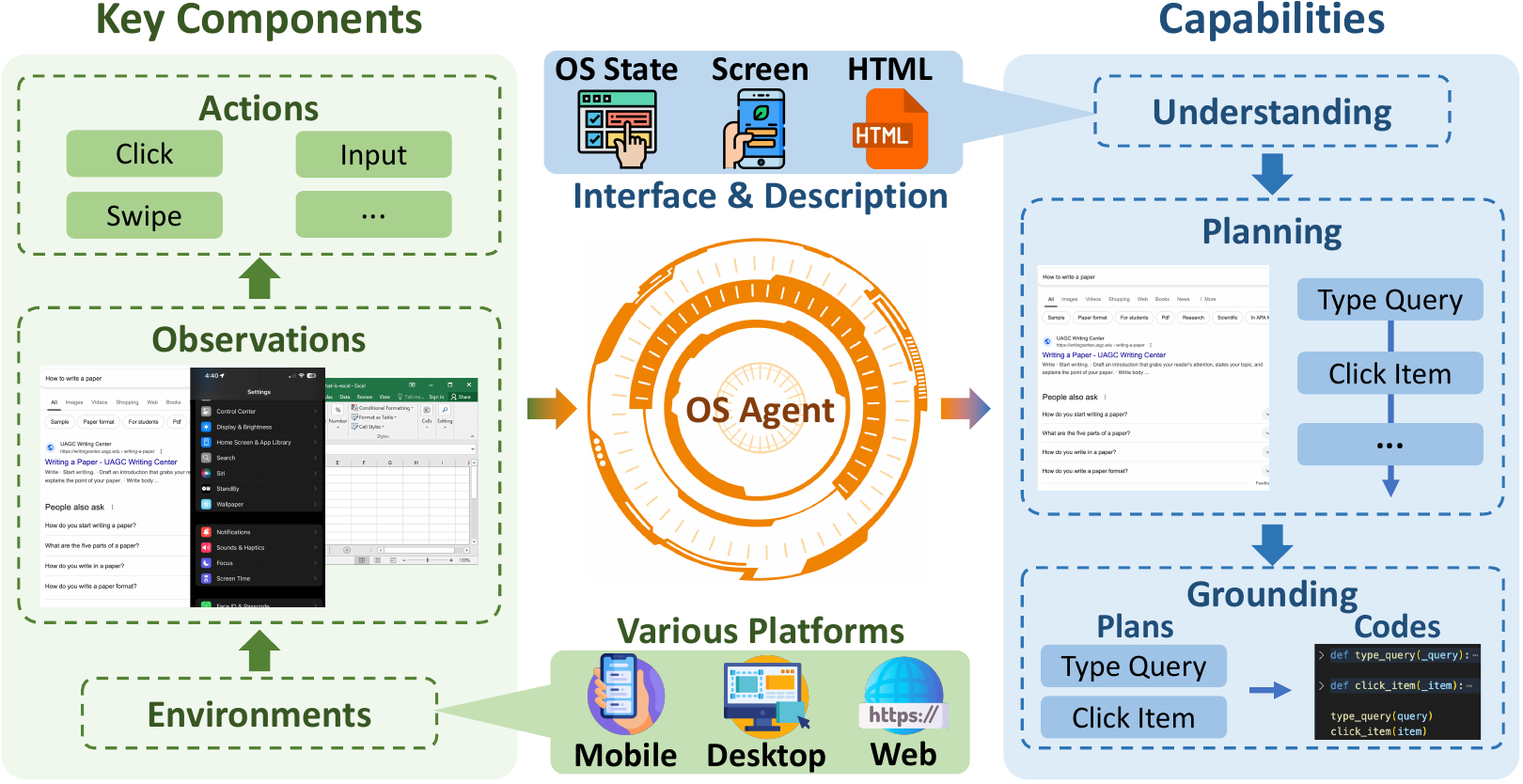}
    \caption{Fundamentals of OS Agents.}
    \label{fig:enter-label}
\end{figure*}

\section{Introduction}
Building a superintelligent AI assistant akin to \textit{J.A.R.V.I.S.}\footnote{J.A.R.V.I.S. stands for ``Just A Rather Very Intelligent System'', a fictional AI assistant character from the Marvel Cinematic Universe. It appears in Iron Man (2008), The Avengers (2012), and other films, serving as Tony Stark’s (Iron Man's) personal assistant and interface for his technology.} from the Marvel movie \textit{Iron Man}, which assists \textit{Tony Stark} in controlling various systems and automating tasks, has long been a human aspiration. These entities are recognized as \textbf{Operating System Agents (OS Agents)}, as they use computing devices (e.g., computers and mobile phones) by operating within the environments and interfaces (e.g., Graphical User Interface (GUI)) provided by operating systems (OS). OS Agents can complete tasks autonomously and have the potential to significantly enhance the lives of billions of users worldwide. Imagine a world where tasks such as online shopping, travel arrangements booking, and other daily activities could be seamlessly performed by these agents, thereby substantially increasing efficiency and productivity. In the past, virtual assistants such as Siri \citep{apple_siri}, Cortana \citep{microsoft_cortana}, Amazon Alexa \citep{google_assistant} and Google Assistant\citep{amazon_alexa} have already offered glimpses into this potential, but limitations in model capabilities such as contextual understanding \citep{tulshan2019survey}, have prevented these products from achieving widespread adoption and full functionality.

Fortunately, recent advancements in (multimodal) large language models ((M)LLMs), such as Gemini \citep{google_gemini}, GPT \citep{openai}, Grok \citep{x_ai}, Yi \citep{lingyiwanwu} and Claude \citep{anthropic} series\footnote{Rankings were determined using the Chatbot Arena LLM Leaderboard \citep{chiang2024chatbot} as of December 12, 2024. For models originating from the same producer, rankings were assigned based on the performance of the highest-ranking model.} have ushered in a new era of possibilities for OS Agents. 
These models boast remarkable abilities, enabling OS Agents to better understand complex tasks and use computing devices to execute. Notable examples include the recently released Computer Use by Anthropic \citep{anthropic_computer_use}, Apple Intelligence by Apple \citep{apple_intelligence}, AutoGLM by Zhipu AI \citep{liu2024autoglm} and Project Mariner by Google Deepmind \citep{deepmind_mariner}. For instance, Computer Use leverages Claude \citep{anthropic_claude_model} to interact directly with users' computers, aiming for seamless task automation. Meanwhile, in the research community, a variety of works have been proposed to build (M)LLM-based OS Agents \citep{gur2023real,you2025ferret,gou2024navigating,meng2024vga,chen2024edge,wu2024atlas,zhang2023appagent,yan2023gpt,ma2023laser,zhang2024android,he2024webvoyager,wang2024oscar}. For instance, \citet{wu2024atlas} proposes OS-Atlas, a foundational GUI action model that significantly improves GUI grounding and Out-Of-Distribution task performance by synthesizing GUI grounding data across various platforms. OS-Copilot \citep{wu2024copilot} is an agent framework crafted to develop generalist agents that automate broad computer tasks, demonstrating robust generalization and self-improvement across diverse applications with minimal supervision. Given these advancements and the growing body of work, it has become increasingly important to provide a comprehensive survey that consolidates the current state of research in this area.

In this survey, we begin by discussing the fundamentals of OS Agents (\S \ref{sec:fundamentals}), starting with a definition of what constitutes an OS Agent. As illustrated in Figure \ref{fig:enter-label}, we focus on three key components: the environment, the observation space, and the action space (\S \ref{sec:key_components}). We then outline the essential capabilities OS Agents should possess, including understanding, planning, and grounding (\S \ref{sec:capabilities}). Next, we explore two critical aspects of constructing OS Agents (\S \ref{sec:construction}): (1) the development of domain-specific foundation models, covering areas such as architectural design, pre-training, supervised fine-tuning, and reinforcement learning (\S \ref{sec:foundation_models}); and (2) the building of effective agent frameworks around these models, addressing core elements including perception, planning, memory, and action (\S \ref{sec:agent_frameworks}). We also review the evaluation protocol (\S \ref{sec:evaluation_protocol}) and benchmarks (\S \ref{sec:evaluation_benchmark}) commonly used to assess the performance of OS Agents. Finally, we discuss the challenges and future directions for OS Agents (\S \ref{sec:challenges_and_future}), with a particular focus on issues related to safety and privacy (\S \ref{sec:safety_and_privacy}), as well as personalization and self-evolution (\S \ref{sec:personalization_and_self_evolution}).

This survey aims to make contributions to the research and development of OS Agents by providing readers with a comprehensive understanding of their essential capabilities, offering insights into methodologies for building OS Agents based on (M)LLMs, and highlighting the latest research trends, challenges and future in this field. Recognizing that OS Agents are still in their early stages of development, we acknowledge the rapid advancements that continue to introduce novel methodologies and applications. To support ongoing developments, we maintain an open-source GitHub repository as a dynamic resource. Through this work, we aspire to inspire further innovation, driving progress in both academic research and industrial applications of OS Agents.

\section{Fundamental of OS Agents}
\label{sec:fundamentals}
OS Agents are specialized AI agents that leverage the environment, input, and output interfaces provided by the operating system to generally using computing devices in response to user-defined goals. These agents are designed to automate tasks executed within the operating system, leveraging the exceptional understanding and generative capabilities of (M)LLMs to enhance user experience and operational efficiency. To achieve this, OS Agents are based on three key components: Environment, Observation Space, and Action Space, which together facilitate the agent's effective engagement with the operating system. Additionally, OS Agents necessitate three core capabilities: Understanding, Planning, and Grounding. These capabilities enable them to sequentially comprehend tasks, devise action strategies, and implement these actions effectively within the environment.

\subsection{Key Component}
\label{sec:key_components}
\textbf{Environment.} 
The environment for OS Agents refers to the system or platform in which they operate. This can include desktop \citep{gao2023assistgui,bonatti2024windows,kapoor2024omniact}, mobile \citep{venkatesh2022ugif,rawles2024androidworld,li2024effects,bishop2024latent,xing2024understanding} or web \citep{shi2017world,yao2022webshop,koh2024visualwebarena,lu2024weblinx,drouin2024workarena,lee2024benchmarking}. OS Agents interact with these diverse environments to perform tasks, gather feedback, and adapt to their unique characteristics. These environments encompass a diverse set of tasks, ranging from simple interactions such as information retrieval to complex multi-step operations, requiring agents to perform planning and reasoning across multiple interfaces, significantly increasing the complexity and posing challenges for OS Agents. We refer readers to \S \ref{sec:evaluation_benchmark} for detailed discussion.

\textbf{Observation Space.}
The observation space encompasses the information OS Agents can access about the system's state and user activities. These observations guide the agents in comprehending the environment, making informed decisions, and determining the appropriate actions to achieve user-defined goals. Observation includes capturing outputs from the OS, such as screen images \citep{yan2023gpt,zhang2023you,zhang2024android,hoscilowicz2024clickagent} with specific processing \citep{zhang2023appagent,he2024webvoyager,fu2024periguru}, or textual data, such as the description of the screen \citep{gao2023assistgui,wu2024copilot} and the HTML code \citep{ma2023laser,zheng2024gpt} in web-based contexts. Multimodal input integrating these diverse data structure introduces significant challenges for agents to effectively understand and execute tasks. Further details are elaborated in \S \ref{sec:perception}.

\textbf{Action Space.}
The action space defines the set of interactions through which OS Agents manipulate the environment using the input interfaces provided by the operating system. These actions can be broadly categorized into input operations \citep{sun2022meta,zhang2023appagent,gao2023assistgui}, representing the primary methods of interacting with digital interfaces, navigation operations \citep{yan2023gpt,song2024beyond,he2024openwebvoyager} which facilitate movement across the system’s interface and extended operations, such as utilizing external tools or services \citep{wu2024copilot,mei2024aios}. These actions enable OS Agents to execute tasks, control applications, and automate workflows effectively. A comprehensive discussion can be found in \S \ref{sec:action}.

\subsection{Capability}
\label{sec:capabilities}
\textbf{Understanding.}
A crucial capability of OS Agents is their ability to comprehend complex OS environments. These environments encompass a diverse array of data formats, including HTML code \citep{gur2023real,lai2024autowebglm} and graphical user interfaces captured in screenshots \citep{nong2024mobileflow,wu2024atlas}. The complexity escalates with length code with sparse information, high-resolution interfaces cluttered with minuscule icons, small text, and densely packed elements \citep{he2024webvoyager,hong2024cogagent,you2025ferret}. Such environments challenge the agents' perceptual abilities and demand advanced contextual comprehension. This comprehension is essential not only for tasks aimed at information retrieval \citep{rawles2024androidworld} but also serves as a fundamental prerequisite for effectively executing a broad spectrum of additional tasks.

\textbf{Planning.}
Planning \citep{huang2023reasoninglargelanguagemodels,zhang2024llmmastermindsurveystrategic,huang2024understanding} is a fundamental capability of OS Agents, enabling them to decompose complex tasks into manageable sub-tasks and devise sequences of actions to achieve specific goals \citep{wu2024copilot,gao2023assistgui}. Planning within operating systems often requires agents to dynamically adjust plans based on environmental feedback and historical actions \citep{zhang2023you,wang2024oscar,kim2024language}. Reasoning strategies like ReAct \citep{yao2023reactsynergizingreasoningacting} and CoAT \citep{zhang2024android} are also necessary to ensure effective task execution in dynamic and unpredictable scenarios.

\textbf{Grounding.}
Action grounding is another essential capability of OS Agents, referring to the ability to translate textual instructions or plans into executable actions within the operating environment \citep{zheng2024gpt,wu2024atlas}. The agent must identify elements on the screen and provide the necessary parameters (e.g., coordinates, input values) to ensure successful execution. While OS environments often contain numerous selectable elements and possible actions, the resulting complexity makes grounding tasks particularly challenging.

\section{Construction of OS Agents}
\label{sec:construction}

In this section, we discuss effective strategies for constructing OS Agents. We begin by focusing on the development of foundation models tailored for OS Agents. Domain-specific foundation models \citep{roziere2023code,wu2023bloomberggpt,singhal2023towards,xiao2021lawformer} can significantly enhance the performance of OS Agents by incorporating specialized knowledge and capabilities essential for interacting with operating systems. This can be achieved through thoughtful model architecture design and targeted training strategies that align with specific tasks in this domain.
In addition, we explore the construction of agent frameworks \citep{Chase_LangChain_2022,Significant_Gravitas_AutoGPT,hong2024metagptmetaprogrammingmultiagent,hu2024infiagent} that build upon these foundation models using non-tuning strategies. Techniques such as reasoning strategies and memory augmentation enable agents to accurately perceive their environment, generate effective plans, and execute precise actions without the need for fine-tuning. These approaches offer flexibility and efficiency, allowing OS Agents to generalize across diverse tasks and environments.
By combining robust domain-specific foundation models with agent frameworks, we can further enhance the adaptability, reliability, and efficiency of OS Agents in automating complex tasks.

\begin{table*}
\centering
\caption{Recent foundation models for OS Agents. Arch.: Architecture, Exist.: Existing, Mod.: Modified, Concat.: Concatenated, PT: Pre-Train, SFT: Supervised Fine-Tune, RL: Reinforcement Learning.} \label{tab:foundation}
\begin{tblr}{
  column{even} = {c},
  column{3} = {c},
  column{5} = {c},
  hline{1,32} = {-}{0.08em},
  hline{2} = {-}{0.05em},
}
\textbf{Model}               & \textbf{Arch.}           &\textbf{PT}  & \textbf{SFT} & \textbf{RL} & \textbf{Date}    \\
OS-Atlas \citep{wu2024atlas}           & Exist. MLLMs          & \checkmark        & \checkmark        & -  & 10/2024 \\
AutoGLM \citep{liu2024autoglm}           & Exist. LLMs           & \checkmark        & \checkmark        & \checkmark  & 10/2024 \\
EDGE \citep{chen2024edge}            & Exist. MLLMs          & -        & \checkmark        & -  & 10/2024 \\
Ferret-UI 2 \citep{li2024ferret}        & Exist. MLLMs          & -        & \checkmark        & -  & 10/2024 \\
ShowUI \citep{lin2024showui}             & Exist. MLLMs          & \checkmark        & \checkmark        & -  & 10/2024 \\
UIX \citep{liu2024harnessing}             & Exist. MLLMs          & -        & \checkmark        & -  & 10/2024 \\
TinyClick \citep{pawlowski2024tinyclick}          & Exist. MLLMs          & \checkmark        & -        & -  & 10/2024 \\
UGround \citep{gou2024navigating}            & Exist. MLLMs          & -        & \checkmark        & -  & 10/2024 \\
NNetNav \citep{murty2024nnetscape}            & Exist. LLMs           & -        & \checkmark        & -  & 10/2024 \\
Synatra \citep{ou2024synatra}            & Exist. LLMs           & -        & \checkmark        & -  & 09/2024 \\
MobileVLM \citep{wu2024mobilevlm}          & Exist. MLLMs          & \checkmark        & \checkmark        & -  & 09/2024 \\
UI-Hawk \citep{zhang2024uihawk}            & Mod. MLLMs      & \checkmark        & \checkmark        & -  & 08/2024 \\
GUI Action Narrator \citep{wu2024gui} & Exist. MLLMs          & -        & \checkmark        & -  & 07/2024 \\
MobileFlow \citep{nong2024mobileflow}         & Mod. MLLMs      & \checkmark        & \checkmark        & -  & 07/2024 \\
VGA \citep{meng2024vga}                & Exist. MLLMs          & -        & \checkmark        & -  & 06/2024 \\
OdysseyAgent \citep{lu2024gui}       & Exist. MLLMs          & -        & \checkmark        & -  & 06/2024 \\
Textual Foresight \citep{burns2024tell}  & Concat. MLLMs  & \checkmark        & \checkmark        & -  & 06/2024 \\
WebAI \citep{thil2024navigating}              & Concat. MLLMs  & -        & \checkmark        & \checkmark  & 05/2024 \\
GLAINTEL \citep{fereidouni2024search}     & Exist. MLLMs           & -        & -        & \checkmark  & 04/2024 \\
Ferret-UI \citep{you2025ferret}          & Exist. MLLMs          & -        & \checkmark        & -  & 04/2024 \\
AutoWebGLM \citep{lai2024autowebglm}         & Exist. LLMs           & -        & \checkmark        & \checkmark  & 04/2024 \\
\citet{patel2024large}                 & Exist. LLMs           & -        & \checkmark        & -  & 03/2024 \\
ScreenAI \citep{baechler2024screenai}           & Exist. MLLMs          & \checkmark        & \checkmark        & -  & 02/2024 \\
Dual-VCR \citep{kil2024dual}           & Concat. MLLMs  & -        & \checkmark        & -  & 02/2024 \\
SeeClick \citep{cheng2024seeclick}           & Exist. MLLMs          & \checkmark        & \checkmark        & -  & 01/2024 \\
CogAgent \citep{hong2024cogagent}           & Mod. MLLMs      & \checkmark        & \checkmark        & -  & 12/2023 \\
ILuvUI \citep{jiang2023iluvui}             & Mod. MLLMs      & -        & \checkmark        & -  & 10/2023 \\
RUIG \citep{zhang2023reinforced}                & Concat. MLLMs  & -        & -        & \checkmark  & 10/2023 \\
WebAgent \citep{iong2024openwebagent}           & Concat. LLMs   & \checkmark        & \checkmark        & -  & 07/2023 \\
WebGUM \citep{furuta2023multimodal}             & Concat. MLLMs  & -        & \checkmark        & -  & 05/2023 
\end{tblr}
\end{table*}

\subsection{Foundation Model} \label{sec:foundation_models}

\begin{figure}
    \centering
    \includegraphics[width=1\linewidth]{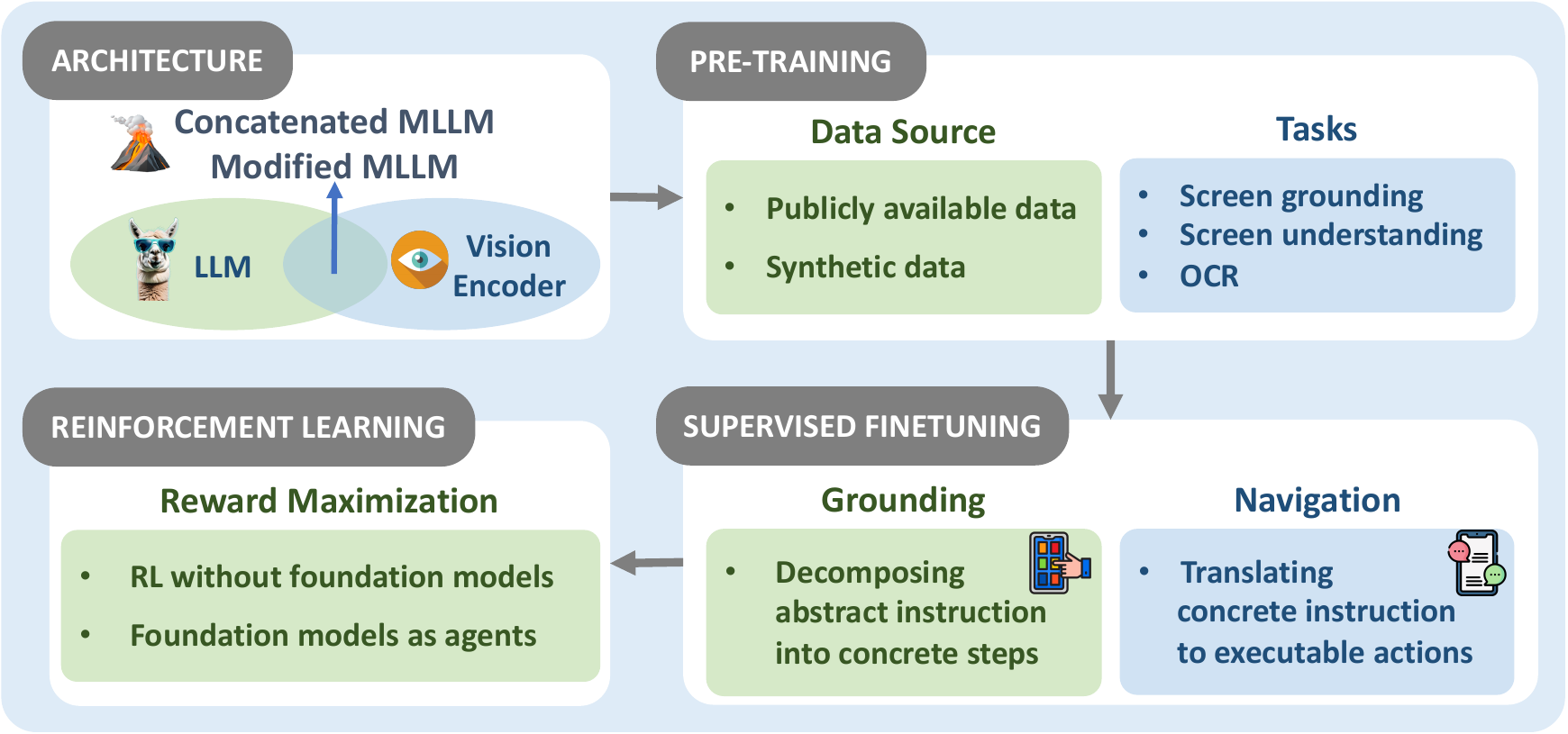}
    \caption{Summary of the content about foundation models for OS Agents in \S\ref{sec:foundation_models}.}
    \label{fig:enter-label}
\end{figure}

The construction of foundation models for OS Agents involves two key components: model architecture and training strategies. The architecture defines how models deal with input and output within OS environments, while training strategies enhance models with the ability of completing complex tasks. As illustrated in Figure \ref{fig:enter-label}, training strategies that are applied in construction of foundation models for OS Agents mainly include pre-training, supervised finetuning and reinforcement learning. Table \ref{tab:foundation} summarizes the architecture and training strategies used in the recent foundation models for OS Agents.

\subsubsection{Architecture}

A variety of architectures are employed to construct foundation models for OS Agents. It is common practice to build these models by leveraging existing open-source LLMs and MLLMs. Some architectures can be created by concatenating LLMs with vision encoders, enabling the models to process both textual and visual information. Additionally, MLLMs are frequently adapted by incorporating supplementary modules to address the specific requirements such as high-resolution image understanding.

\textbf{Existing LLMs.} The architecture of existing LLMs can already process user instructions and read HTML code to perceive information contained in user interfaces. Therefore, several works \citep{liu2024autoglm, lai2024autowebglm, patel2024large} directly chose open-source LLMs as backbone models without further optimizing architecture to develop foundation models for OS Agents, where T5 \citep{fereidouni2024search, furuta2024exposing} and LLaMA \citep{murty2024nnetscape, ou2024synatra} are popular architectures. WebAgent \citep{gur2023real} combines Flan-U-PaLM with HTML-T5, a finetuned version of Long-T5-base. HTML-T5 reads user instructions together with HTML code of user interface and navigation history to produce a summary of the user interface and a plan for completing tasks specified in the user instruction, which would then be processed by the Flan-U-PaLM instance that generates executable Python code to execute user instructions.

\textbf{Existing MLLMs.} LLMs are capable of handling OS tasks, while an inescapable shortcoming of LLMs is that LLMs can only process textual input, while GUI are designed for human users that directly perceive vision information to operate the apps. For this, MLLMs, which additionally have the ability to process vision information while preserving the ability for complex natural language processing, are introduced. Various works \citep{baechler2024screenai, chen2024edge, pawlowski2024tinyclick} have shown that architectures of existing MLLMs such as LLaVA \citep{gou2024navigating, meng2024vga}, Qwen-VL \citep{cheng2024seeclick, lu2024gui, wu2024gui}, InternVL \citep{wu2024atlas, gao2024mobileviews}, CogVLM \citep{zhang2024android, xu2024androidlab}, etc., can be effective for developing foundation models for OS Agents.

\textbf{Concatenated MLLMs.} Typical architecture of MLLMs consists of an LLM and a vision encoder connected by an adapter network or a cross-attention module. Several works \citep{kil2024dual, zhang2023reinforced} have shown that choosing LLMs and vision encoders that are suitable to process OS tasks and concatenating them in a way that is similar to that of existing MLLMs' could be a more suitable approach for constructing foundation models for OS Agents. For instance, \citet{furuta2023multimodal} and \citet{thil2024navigating} chose T5 as the LLM in the structure, whose encoder-decoder architecture can better fit tree-architecture of HTML, enabling the model to better process GUI information by perceiving both text and image forms of the GUI. 

\textbf{Modified MLLMs.} Further adjustments have been adopted to architectures of MLLMs to enhance understanding abilities of foundation models. For instance, most existing MLLMs can only process images of relatively low resolutions, typically 224×224, while common resolution of GUI screenshots is 720×1080. Resizing screenshots to fit the resolution vision encoders of MLLMs preserves features of general layout and most objects, but text and small icons cannot be well perceived, which sometimes would be vital for MLLMs to accomplish OS tasks. Some works have been proposed to enable MLLMs to perceive these features. CogAgent \citep{hong2024cogagent} introduced additional EVA-CLIP-L high-resolution vision encoder that accepts images of size 1120×1120, and added a cross-attention module to connect with the original MLLM. Ferret-UI \citep{you2025ferret} applied the idea of any-resolution, where screenshot images are both resized to fit the vision encoder and partitioned into sub-images, enabling the model to perceive and process visual features in all granularities. MobileFlow \citep{nong2024mobileflow} chose Qwen-VL as the backbone with a GUI encoder (LayoutLMv3) added to the original architecture, which extracts embeddings of both images and OCR texts together with their positions. UI-Hawk \citep{zhang2024uihawk} uses a vision encoder that applies a shape-adaptive cropping strategy to perceive details in the screenshot.

\subsubsection{Pre-training}
\label{sec:pre-training}
Pre-training~\citep{devlin2018bert, brown2020language, dosovitskiy2020image} lays the foundation for model construction and is extensively employed to enhance the foundation models for OS Agents by expanding their understanding of GUI and facilitating the acquisition of the inherent correlations between visual and textual information. To achieve this, most existing pre-training approaches utilize continual pre-training from general pre-trained models with substantial textual or visual comprehension capabilities. This strategy leverages the established knowledge within these pre-trained models, thereby enhancing their performance on GUI-related tasks. One exception is \citet{gur2023real}, who trained their model from scratch, focusing specifically on parsing HTML text without incorporating the visual modality. To provide a comprehensive overview of their impact on the development of foundation models for OS Agents, data sources and tasks in pre-training will be discussed in the following.

\textbf{Data source.}
(1) \textit{Publicly available data.} Some studies leverage publicly available datasets to quickly obtain large-scale data for pre-training. Specifically, \citet{gur2023real} crawled and filtered web data to extract GUI-related information. \citet{gur2023real} utilized CommonCrawl to acquire HTML documents, removing those with non-unicode or purely alphanumeric content, and extracted subtrees around `<label>' elements to train HTML-T5, a model capable of providing executable instructions. Similarly, \citet{nong2024mobileflow} employed Flickr30K for modality alignment, enhancing the model's semantic understanding of images. However, relying solely on publicly available data for pre-training is insufficient to address the complex and diverse tasks required by OS Agents~\citep{gou2024navigating}. Consequently, (2) \textit{Synthetic data.} Researchers incorporate synthetic data into the pre-training process, inspired by the real-world application scenarios of OS Agents. 
\citet{cheng2024seeclick} extracts visible text element positions and instructions to build grounding\footnote{Given the varying interpretations of 'grounding' across different domains, in this subsubsection, the term 'grounding' specifically refers to visual grounding, which is the process of locating objects or regions in an image based on a natural language query. This definition differs from the one used in \S \ref{sec:capabilities}.} and OCR task data based on HTML data obtained from the web, while \citet{chen2024guicourse} rendered entire websites after acquiring webpage links, segmented them into 1920×1080 resolution screenshots, and extracted features, thereby enriching the diversity of web data. Some studies \citep{wu2024atlas} have noted that although similarities exist between different GUI platforms, pre-training solely based on web data struggles to generalize across platforms. To address this, they created multiple simulated environments and utilized accessibility (A11y) trees to simulate human-computer interaction, sampling cross-platform grounding data. Additionally, \citet{wu2024mobilevlm} proposed a data collection algorithm that simulates human interaction with smartphones by iteratively interacting with every element on each GUI page. This process represents the results as directed graphs and yielded a dataset containing over 3 million real GUI interaction samples.

\textbf{Task.}
(1) \textit{Screen grounding.} Many studies have demonstrated that pre-training enables models to extract 2D coordinates or bounding boxes of target elements from images based on textual descriptions \citep{wu2024atlas, baechler2024screenai, pawlowski2024tinyclick, hong2024cogagent, wu2024mobilevlm, chen2024guicourse, zhang2024uihawk, lin2024showui}. In addition, \citet{cheng2024seeclick, lin2024showui} extended text-based grounding tasks by incorporating requirements for predicting text from center point coordinates and bounding boxes into the pre-training stage. (2) \textit{Screen understanding.} Several studies posit that the foundation models for OS Agents should be capable of extracting semantic information from images, as well as analyzing and interpreting the entire content of the image. \citet{wu2024atlas} emphasized that pre-training should equip MLLMs with the knowledge to understand GUI screenshots and identify elements on the screen. Furthermore, \citet{baechler2024screenai, zhang2024uihawk} proposed screen question-answering as a task, where the former designed datasets targeting tasks involving counting, arithmetic operations, and interpreting complex data in charts. (3) \textit{Optical Character Recognition (OCR).} OCR plays a crucial role in handling GUI elements that contain textual content. \citet{hong2024cogagent} constructed training data during the pre-training stage by using Paddle-OCR to extract text and bounding boxes from GUI screenshots, and validated the model's superior OCR capabilities on the TextVQA benchmark. \citet{lin2024showui} identified the capabilities of OCR as a critical evaluation criterion for constructing foundation models.

\subsubsection{Supervised Finetuning}

Supervised Finetuning (SFT) has been widely adopted to enhance the planning and grounding capabilities of OS Agents. 
This requires efforts to collect domain-specific data to bridge the domain gap between tasks on natural images and GUIs~\citep{hong2024cogagent}, which is thus the key challenge herein.

For planning, researchers first collect multi-step trajectories and synthesize instructions for them.
\citet{gao2024mobileviews} traverse across the apps with fixed rules as well as LLMs, where the latter are applied to handle certain predefined scenarios and cases that fixed rules fail to cover. 
\citet{ou2024synatra} uses online tutorial articles to build trajectories, where descriptions of steps are mapped into agent actions with LLMs.
\citet{chen2024webvln} builds directed graphs about navigation among webpages and finds the shortest path in the graph to obtain trajectories when generating data for certain tasks.
These trajectories are taken into advanced large language models, such as GPT4, to synthesize corresponding task instructions~\citep{hong2024cogagent,you2025ferret} as well as Chain-of-Thought reasoning process to decompose the tasks~\citep{lai2024autowebglm}.

To synthesize data for grounding ability, researchers first connect the actions on the objects to GUI images and then synthesize instructions referring to them.
Common strategies to draw the connections are rendering the source codes of GUIs.
For example, \citet{gou2024navigating,chen2024edge,liu2024harnessing,kil2024dual} render webpages with HTML and \citet{wu2024atlas,baechler2024screenai,gao2024mobileviews,you2025ferret} leverage desktop or mobile simulators.
A few attempts also leverage GUI detection models~\citep{you2025ferret,zhang2024android}.
Compared to simply learning to operate on the source code, learning to operate with their visual form can show superior performance with the straightforward interaction between widgets~\citep{kil2024dual}.
Meanwhile, \citet{meng2024vga} shows learning with GUI images helps avoid hallucination and \citet{liu2024harnessing} demonstrates generalization to unseen GUIs. 
Then, to synthesize instruction referring to the widgets, \citet{gou2024navigating} summarizes three typical expressions, namely referring to their salient visual features, locations, or functions.
Notably, different GUIs may involve different action spaces, \citet{wu2024atlas} find it necessary to adapt action sequences from different sources to a unified action space so as to avoid conflict among them during fine-tuning.

\subsubsection{Reinforcement Learning}

Reinforcement learning (RL) \citep{sutton2018reinforcement} is a machine learning paradigm where agents learn optimal decision-making through interactions with an environment. By receiving feedback in the form of rewards, the agent iteratively refines its strategies to maximize cumulative rewards. 

Early attempts \citep{zheran2018reinforcement,shi2017world,gur2018learning,jia2019dom,shvo2021appbuddy} utilized RL to train agents to accomplish tasks on web and mobile Apps. 
We introduce several representative works as follows. \citet{yao2022webshop} introduced WebShop, a simulated e-commerce website environment, based on which they trained and evaluated a diverse range of agents using reinforcement learning, imitation learning, and pre-trained multimodal models.
The reward is determined by how closely the purchased product matches the specific attributes and options mentioned in the user instructions.
Reinforcement learning is typically combined with behavior cloning or supervised fine-tuning to enhance performance.
For example,
\citet{humphreys2022data} developed a scalable method using reinforcement learning and behavioral priors from human-computer interactions to control computers via keyboard and mouse, achieving human-level performance in the MiniWob++ benchmark.
\citet{zhang2023reinforced} developed a multimodal model for automating GUI tasks by grounding natural language instructions to GUI screenshots, using a pre-trained visual encoder and language decoder, with RL to enhance spatial decoding by supervising token sequences with visually semantic metrics. 

In the above RL-based works, large models generally function as feature extractors. More recently, research has progressed to the ``LLMs as agents'' paradigm, where LLMs serve as policy models and reinforcement learning is applied to align the large models with the final objectives.
\citet{thil2024navigating} improved web navigation in LLMs using the Miniwob++ benchmark by fine-tuning T5-based models with hierarchical planning and then integrating these with a multimodal neural network, utilizing both supervised and reinforcement learning.
\citet{fereidouni2024search} employs the Flan-T5
architecture and introduce training via Reinforcement Learning. They leveraged human demonstrations through behavior cloning and then further trained the agent with PPO.
\citet{liu2024autoglm} followed the paradigm of LLMs as agents and proposed AutoGLM, foundation agents for autonomous control of computing devices through GUIs. They designed an intermediate interface that
effectively disentangles planning and grounding behaviors, and developed a self-evolving online curriculum RL approach that enables robust error recovery and performance improvement. 
\citet{heagile} introduced a novel RL framework for LLM-based 
Agents, AGILE, integrating LLMs, memory, tools, and executor modules. RL enables LLMs to predict actions and the executor to manage them, enhancing decision-making and interactions. Reinforcement learning is also introduced to the agents based on vision-only models \citep{shaw2023pixels} and MLLMs \citep{bai2024digirl,wang2024distrl}.

\begin{table}
\caption{Recent agent frameworks for OS Agents. TD: Textual Description, GS: GUI Screenshots, VG: Visual Grounding, SG: Semantic Grounding, DG: Dual Grounding, GL: Global, IT: Iterative, AE: Automated Exploration, EA: Experience-Augmented, MA: Management, IO: Input Operations, NO: Navigation Operations, EO: Extended Operations.}
\label{tab:frameworks}
\centering
\scalebox{.85}{
\begin{tblr}{
  column{even} = {c},
  column{3} = {c},
  column{5} = {c},
  hline{1,33} = {-}{0.08em},
  hline{2} = {-}{0.05em},
}
\textbf{Agent}          & \textbf{Perception} & \textbf{Planning} & \textbf{Memory}            & \textbf{Action}~  & \textbf{Date}    \\
OpenWebVoyager \citep{he2024openwebvoyager} & GS, SG      & -      & -            & IO, NO    & 10/2024 \\
OSCAR \citep{wang2024oscar}         & GS, DG      & IT           & AE           & EO       & 10/2024 \\
PUMA \citep{cai2024large}           & TD         & -           & -            & IO, NO, EO & 10/2024 \\
AgentOccam \citep{yang2024agentoccam}     & TD         & IT     & MA           & IO, NO    & 10/2024 \\
Agent S \citep{agashe2024agent}       & GS, SG      & GL      & EA, AE, MA     & IO, NO    & 10/2024 \\
ClickAgent \citep{hoscilowicz2024clickagent}    & GS         & IT        & AE           & IO, NO    & 10/2024 \\
LSFS \citep{shi2024commands}          & GS, SG      & -        & -             & EO       & 09/2024 \\
NaviQAte \citep{shahbandeh2024naviqatefunctionalityguidedwebapplication}      & GS, SG           & -      & -            & IO       & 09/2024 \\
PeriGuru \citep{fu2024periguru}       & GS, DG      & IT          & EA, AE        & IO, NO    & 09/2024 \\
OpenWebAgent \citep{iong2024openwebagent}   & GS, DG      & -           & -            & IO       & 08/2024 \\
LLMCI  \citep{barham2024towards}        & GS, SG      & -            & -            & EO       & 07/2024 \\
Agent-E \citep{abuelsaad2024agent}        & TD         & IT           & AE, MA        & IO, NO    & 07/2024 \\
Cradle \citep{tan2024cradle}         & GS         & IT         & EA, AE, MA     & EO       & 03/2024 \\
CoAT  \citep{zhang2024android}        & GS         & IT          & -            & IO, NO    & 03/2024 \\
Self-MAP \citep{deng2024multi}      & -          & IT          & EA           & IO       & 02/2024 \\
OS-Copilot \citep{wu2024copilot}    & TD         & GL           & EA, AE        & IO, EO    & 02/2024 \\
Mobile-Agent \citep{wang2024mobile}   & GS, SG      & IT          & AE           & IO, NO    & 01/2024 \\
WebVoyager \citep{he2024webvoyager}    & GS, VG      & IT         & MA           & IO, NO    & 01/2024 \\
AIA \citep{ding2024mobileagent}           & GS, VG      & GL             & -            & IO, NO    & 01/2024 \\
SeeAct \citep{zheng2024gpt}        & GS, SG      & -         & AE           & IO       & 01/2024 \\
AppAgent \citep{zhang2023appagent}      & GS, DG      & IT          & AE           & IO, NO    & 12/2023 \\
ACE \citep{gao2023assistgui}           & TD         & GL        & AE           & IO, NO    & 12/2023 \\
MobileGPT \citep{lee2023explore}     & TD         & GL         & MA           & IO, NO    & 12/2023 \\
MM-Navigator \citep{yan2023gpt}  & GS, VG      & -          & MA           & IO, NO    & 11/2023 \\
WebWise \citep{tao2023webwise}       & TD         & -             & MA           & IO, NO    & 10/2023 \\
\citet{li2023zero}      & TD         & IT          & AE           & IO, NO    & 10/2023 \\
Laser \citep{ma2023laser}         & TD         & IT           & AE           & IO, NO    & 09/2023 \\
Synapse \citep{zheng2023synapse}       & -          & -            & MA           & IO       & 06/2023 \\
SheetCopilot \citep{li2024sheetcopilot}  & TD         & IT       & AE           & EO       & 05/2023 \\
RCI \citep{kim2024language}           & -         & IT           & AE           & IO, NO    & 03/2023 \\
\citet{wang2023enabling}    & TD              & -      & -            & IO       & 09/2022 \\
\end{tblr}
}
\end{table}

\subsection{Agent Framework}
\label{sec:agent_frameworks}
\begin{figure}
    \centering
    \includegraphics[width=1\linewidth]{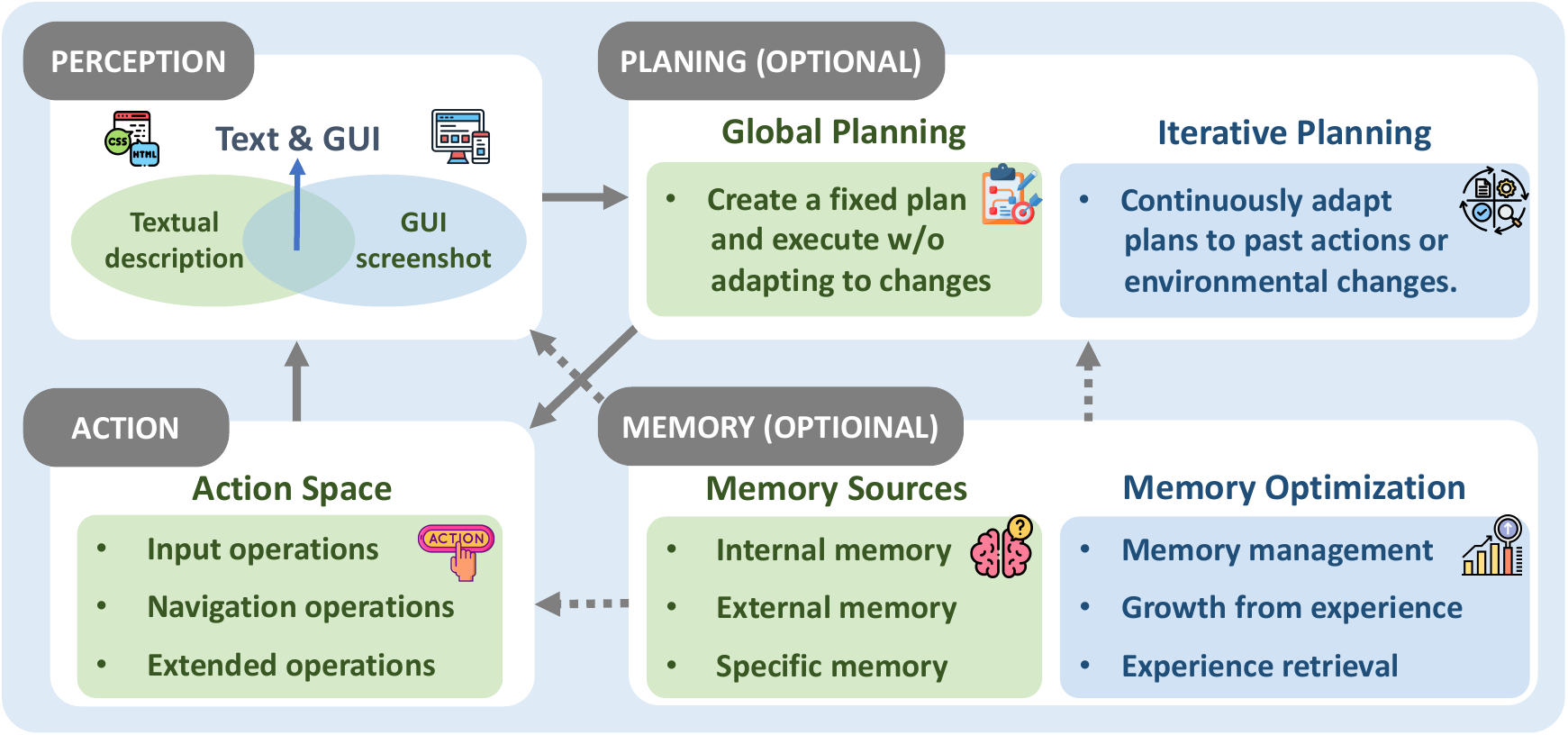}
    \caption{Summary of the content about agent frameworks for OS Agents in \S\ref{sec:agent_frameworks}.}
    \label{fig:frameworks}
\end{figure}

OS Agent frameworks typically consist of four core components: Perception, Planning, Memory, and Action. The perception module collects and analyzes environmental information; the planning module handles task decomposition and action sequence generation; the memory module supports information storage and experience accumulation; and the action module executes specific operation instructions. As illustrated in Figure \ref{fig:frameworks}, these components work together to enable OS Agents to understand, plan, remember, and interact with operating systems. Table \ref{tab:frameworks} summarizes the technical characteristics of recent OS Agent frameworks, including their specific implementations across these four core components. 

\subsubsection{Perception}
\label{sec:perception}
Perception is the process through which OS Agents collect and analyze information from their environment. 
In OS Agents, the perception component needs to observe the current environment and extract relevant information to assist with the agents' planning, action, and memory optimization.
Perception can be broadly categorized into two types based on the input modality as follows:

\textbf{Textual Description of OS.}
Early works \citep{ma2023laser,wang2023enabling,lee2023explore,gao2023assistgui,li2024sheetcopilot,wu2024copilot,lu2024turn} are limited by the fact that LLMs could only process textual input. Therefore, they mainly rely on using tools to convert OS states into text descriptions.

To facilitate LLMs' understanding, these text descriptions are often represented in a structured format, such as HTML, DOM, or accessibility tree.
For instance, MobileGPT \citep{lee2023explore} converts mobile screens into a simplified HTML representation to help LLMs' comprehension.
However, these approaches may generate irrelevant or redundant information, which can negatively impact the OS Agents' judgment of the environment and lead to incorrect actions.
Therefore, some new approaches have been proposed to filter out invalid descriptions, ensuring that OS Agents only observe relevant information.
For example, Agent-E \citep{abuelsaad2024agent} introduces a flexible DOM distillation approach that allows the agent to choose the most suitable DOM representation from three different implementations based on the specific task at hand.
\citet{li2023zero} only expands the HTML representation when the agent takes action, compelling it to make rational decisions with limited information.
WebWise \citep{tao2023webwise} introduces a filtering function \textit{filterDOM} to select relevant DOM elements based on predefined ``tags'' and ``classes,'' filtering out unnecessary items.

\textbf{GUI Screenshot.}
The emergence of MLLMs enables OS Agents to process visual inputs.
Research is increasingly treating GUI screenshots as the perception input for OS Agents, which better aligns with human behavior.
However, most existing vision encoders of OS Agents are pre-trained on general data, which makes OS Agents less sensitive to GUI elements.
To enhance OS Agents' understanding ability of GUI screenshots without fine-tuning visual encoders, existing research focuses on GUI grounding\footnote{Given the varying interpretations of `grounding' across different domains, in this subsubsection, the term `grounding' specifically refers to GUI grounding, which is the process of enhancing OS Agents' understanding ability of GUI screenshots through prompts. This definition differs from the one used in \S \ref{sec:capabilities} or \S \ref{sec:pre-training}.}.

GUI grounding refers to the process of interpreting the screen and accurately identifying relevant elements, such as buttons, menus, or text fields. GUI grounding can generally be categorized into three types:
(1) \textit{Visual grounding.} Most research \citep{yan2023gpt,wang2024mobile} uses SoM prompting \citep{yang2023setofmarkpromptingunleashesextraordinary} to enhance OS Agents' visual grounding ability.
They incorporate techniques like OCR and GUI element detection algorithms such as ICONNet \citep{sunkara2022towards} and Grounding DINO \citep{liu2024groundingdinomarryingdino} to extract bounding boxes of interactive elements, which are then integrated into corresponding image regions to enhance agents' understanding of GUI screenshots.
(2) \textit{Semantic grounding.} Some studies improve OS Agents' semantic grounding ability by adding descriptions of these interactive elements.
Specifically, SeeAct \citep{zheng2024gpt} enhances semantic grounding by using the HTML document of a website as the semantic reference for the GUI screenshot, thereby linking the visual elements with their corresponding semantic meaning in the HTML structure.
(3) \textit{Dual grounding.} Dual grounding combines both visual and semantic information to improve OS Agents' understanding of the visual environment.
For instance, AppAgent \citep{zhang2023appagent} inputs a labeled screenshot along with an XML file that details the interactive elements to enhance agent understanding.
OSCAR \citep{wang2024oscar} introduces a dual-grounding observation approach, using a Windows API-generated A11Y tree for GUI component representation and adding descriptive labels for semantic grounding.
PeriGuru \citep{fu2024periguru} inputs a labeled screenshot and a detailed description generated through element and layout recognition. DUAL-VCR \citep{,kil2024dual} employs a Dual-View Contextualized Representation approach, extracting visual features using the Pix2Struct Vision Transformer \citep{lee2023pix2struct} and aligning each element with corresponding ``HTML text'' following MindAct \citep{deng2024mind2web} for semantic grounding.

\subsubsection{Planning}
Planning is the process of developing a sequence of actions to achieve a specific goal based on the current environment \citep{huang2023reasoninglargelanguagemodels,zhang2024llmmastermindsurveystrategic,huang2024understanding}. 
It enables OS Agents to break down complex tasks into smaller, manageable sub-tasks and solve them step by step. 
Unlike general agents, the environment of OS Agents is constantly evolving. For instance, dynamic web pages change over time, and GUIs also adapt after each action is executed. 
Therefore, feasible planning is crucial for OS Agents to effectively cope with these ongoing environmental changes. 
We categorize existing studies into two key approaches based on whether the planning is fixed or iterates in response to environmental changes: global planning and iterative planning, detailed as follows:

\textbf{Global}. OS Agents only generate a global plan once and execute it without making adjustments based on environmental changes.
Chain-of-Thought (CoT) \citep{wei2023chainofthoughtpromptingelicitsreasoning} prompts (M)LLMs to break down complex tasks into reasoning steps, which forms the foundation of global planning in most OS Agents \citep{fu2024periguru}.
Due to the one-time nature of global planning, research on global planning focuses on fitting the OS Agents' environment and tasks, proposing sufficiently feasible plans from the outset. For example, OS-Copilot \citep{wu2024copilot} leverages LLMs to formalize the global plan into a directed acyclic graph, enabling parallel execution of independent sub-tasks, which minimizes execution time and improves efficiency. ACE \citep{gao2023assistgui} prompts LLMs to refine extracted steps in alignment with user queries. Agent S \citep{agashe2024agent} proposes experience-augmented hierarchical planning, where plans are informed by integrating knowledge from memory and online sources. Similarly, AIA \citep{ding2024mobileagent} utilizes Standard Operating Procedures (SOP) to break down complex tasks into manageable sub-tasks.

\textbf{Iterative.} In contrast to global planning, iterative planning allows OS Agents to continuously iterate their plans based on historical actions or changes in the environment, enabling them to adapt to ongoing environmental changes.
This methodology is crucial for OS Agents to handle dynamic and unpredictable environments effectively.
In specific, ReAct \citep{yao2023reactsynergizingreasoningacting} builds on the concept of CoT by integrating reasoning with the outcome of actions, making planning more adaptable to changes in the environment. 
This approach has been widely applied in OS Agents \citep{zhang2023appagent,ma2023laser,he2024webvoyager,hoscilowicz2024clickagent,wang2024mobile} for iterative planning.
Reflexion \citep{shinn2023reflexionlanguageagentsverbal} builds upon ReAct by allowing access to previous actions and states, which enhances strategic planning of OS Agents in complex, time-sensitive scenarios \citep{fu2024periguru,tan2024cradle,abuelsaad2024agent}.
In addition to these general iterative planning methods, some studies have proposed iterative planning approaches specifically tailored for OS Agents.
For instance, Auto-GUI \citep{zhang2023you} employs a CoT technique, where a history of past actions is used to generate future plans iteratively after each step. OSCAR \citep{wang2024oscar} introduces task-driven replanning, allowing the OS Agent to modify its plan based on real-time feedback from the environment. SheetCopilot \citep{li2024sheetcopilot} employs State Machine-based Task Planning, where proposed plans are revised using either a feedback-based mechanism or a retrieval-based approach, enhancing the OS Agent's ability to adapt to dynamic environments. RCI \citep{kim2024language} prompts LLMs to find problems in their output and improve the output based on what they find, assisting the OS Agent in refining its reasoning process, which leads to more effective and accurate planning.
CoAT \citep{zhang2024android} introduces a more complex and OS Agent-targeted reasoning method compared to ReAct. It prompts the LLMs to perform a reasoning process involving Screen Description, Action Thinking, and Next Action Description, ultimately leading to an Action Result.

\subsubsection{Memory}
As the complexity of automated tasks in operating systems continues to increase, enhancing the intelligence and execution efficiency of OS Agents has become a key research focus. Among these studies, the memory module serves as one of the core components. Using memory effectively, OS Agents can continuously optimize their performance during task execution, adapt to dynamic environments, and perform tasks in various complex scenarios. In this section, we discuss current research advancements related to memory in OS Agents.

\textbf{Memory Sources.}
Memory can be categorized into Internal Memory, External Memory, and Specific Memory, each serving distinct functions in task execution: immediate information storage, external knowledge support, and operation optimization, respectively.
In recent years, research has increasingly focused on improving memory adaptability and diversity to meet the demands of more complex tasks~\citep{zhou2023recurrentgpt,deng2024multi,wang2024agent,huang2024promptrpa,kim2024auto}. 
For example, the introduction of dynamic memory management mechanisms optimizes memory retrieval and updates, while the integration of multimodal approaches further broadens the types and scope of memory data, enabling agents to access more diverse information sources when handling complex scenarios.

\begin{itemize}[leftmargin=*]
\item {
    \textbf{Internal Memory}. In the following, we introduce several components of Internal Memory. (1) \textit{Action History.} By recording each step of operations, the action history helps OS Agents track task paths and optimize decisions.  For instance, Auto-GUI \citep{zhang2023you} integrates historical and future action plans through the chain of previous action histories. (2) \textit{Screenshots.} The storage of screenshots supports visual reasoning and the recognition of GUI components. For example, CoAT \citep{zhang2024android} semantically processes screenshots to extract interface information, enabling better understanding of the task scene. \citet{rawles2024androidinthewild,wang2024oscar} utilize screenshots annotated with Set-of-Mark (SoM) to support visual reasoning, accurately identify GUI components, and perform precise operations, while also aiding in task planning and validation. ToL \citep{pointedlayout} uses GUI screenshots as input to construct a Hierarchical Layout Tree and combines visual reasoning to generate descriptions of content and layout. (3) \textit{State Data.} Dynamic information from the environment, such as page positions and window states, are stored to help OS Agents quickly locate task objectives and maintain high task execution accuracy in changing environments. Specifically, CoCo-Agent \citep{ma2024coco} records layouts and dynamic states through Comprehensive Environment Perception (CEP), while 
     \citet{abuelsaad2024agent,tao2023webwise} employ Document Object Model denoising techniques to dynamically store page information. In the following, we present the two forms of internal memory.
    
    Short-term Memory stores immediate information about the current task, including the action history of the agent, state information, and the execution trajectory of the task. It supports decision optimization and task tracking, providing contextual support for the ongoing task. Recent advances focus on improving the memory capabilities of OS Agents. For example, understanding the layout of objects in a scene through visual information enables multimodal agents to possess more comprehensive cognitive abilities when handling complex tasks.

    Long-term Memory stores historical tasks and interaction records, such as the execution paths of previous tasks, providing references and reasoning support for future tasks. For example, OS-Copilot \citep{wu2024copilot} stores user preferences and the agent's historical knowledge, such as semantic knowledge and task history, as declarative memory. This is used to make personalized decisions and execute tasks, while dynamically generating new tools or storing task-related skill codes during task execution \citep{tan2024cradle}.
    }
    
    \item{
    \textbf{External Memory.} External memory provides long-term knowledge support, primarily enriching an agent's memory capabilities through knowledge bases, external documents, and online information. For instance, agents can retrieve domain-specific background information from external knowledge bases to make more informed judgments in tasks requiring domain expertise. Additionally, some agents dynamically acquire external knowledge by invoking tools such as Application Programming Interfaces (APIs) \citep{song2024beyond,reddy2024infogent}, integrating this knowledge into their memory to assist with task execution and decision optimization. 
    }
    \item{
    \textbf{Specific Memory.} Specific memory focuses on storing information directly related to specific tasks and user needs while incorporating extensive task knowledge and optimized application functions, which can be stored internally or extended through external data sources \citep{zhu2024moba}. 
    Specific Memory can store task execution rules, subtask decomposition methods, and domain knowledge \citep{wang2024mobile}. It provides agents with prior knowledge to assist in handling complex tasks. For instance, MobileGPT \citep{lee2023explore} adopts a three-tier hierarchical memory structure (task, sub-task, action) and organizes memory in the form of a transition graph, breaking tasks down into sub-tasks represented as function calls for quick access and efficient invocation, while CoCo-Agent \citep{ma2024coco} employs task decomposition and Conditional Action Prediction (CAP) to store execution rules and methods. In terms of interface element recognition and interaction, \citet{agashe2024agent,wang2024oscar,he2024openwebvoyager} enhance task understanding by parsing the Accessibility Tree to obtain information about all UI elements on the screen. 
    
    Additionally, Specific Memory can also be used to record user profiles, preferences, and interaction histories to support personalized recommendations, demand prediction, and inference of implicit information. For example, OS-Copilot \citep{wu2024copilot} records user preferences through user profiles, such as tool usage habits and music or video preferences, enabling personalized solutions and recommendation services. Moreover, Specific Memory also supports recording application function descriptions and page access history to facilitate cross-application operation optimization and historical task tracking. For instance, AppAgent \citep{zhang2023appagent} learns application functionality by recording operation histories and state changes, storing this information as documentation. Similarly, ClickAgent \citep{hoscilowicz2024clickagent} improves understanding and operational efficiency in application environments by using GUI localization models to identify and locate GUI elements within applications, while also recording functionality descriptions and historical task information.
    }
\end{itemize}

\textbf{Memory Optimization.}
Memory optimization can enhance an agent's efficiency in operations and decision-making during complex tasks by effectively managing and utilizing memory resources. In the following, we introduce several key strategies.
\begin{itemize}[leftmargin=*]

\item { \textbf{Management.}
    For humans, memory information is constantly processed and abstracted in the brain. Similarly, the memory of OS Agents can be effectively managed to generate higher-level information, consolidate redundant content, and remove irrelevant or outdated information. Effective memory management enhances overall performance and prevents efficiency loss caused by information overload. In specific, \citet{yan2023gpt,tan2024cradle} introduce a multimodal self-summarization mechanism, generating concise historical records in natural language to replace directly storing complete screens or action sequences. WebAgent \citep{gur2023real} understands and summarizes long HTML documents through local and global attention mechanisms, as well as long-span denoising objectives. On the other hand, WebVoyager \citep{he2024webvoyager} employs a Context Clipping method, retaining the most recent three observations while keeping a complete record of thoughts and actions from the history. 
However, for longer tasks, this approach may lead to the loss of important information, potentially affecting task completion. Additionally, Agent-E \citep{abuelsaad2024agent} optimizes webpage representations by filtering task-relevant content, compressing DOM structure hierarchies, and retaining key parent-child relationships, thereby reducing redundancy. AGENTOCCAM \citep{yang2024agentoccam} optimizes the agent's workflow memory through a planning tree, treating each new plan as an independent goal and removing historical step information related to previous plans.
}

\item{
    \textbf{Growth Experience.}
By revisiting each step of a task, the agent can analyze successes and failures, identify opportunities for improvement, and avoid repeating mistakes in similar scenarios \citep{kim2024language}. For instance, MobA \citep{zhu2024moba} introduces dual reflection, evaluating task feasibility before execution and reviewing completion status afterward. Additionally, In \citep{li2023zero}, the agent analyzes the sequence of actions after a task failure, identifies the earliest critical missteps, and generates structured recommendations for alternative actions.
OS Agents can return to a previous state and choose an alternative path when the current task path proves infeasible or the results do not meet expectations, which is akin to classic search algorithms, enabling the agent to explore multiple potential solutions and find the optimal path. For example, LASER \citep{ma2023laser} uses a Memory Buffer mechanism to store intermediate results that were not selected during exploration, allowing the agent to backtrack flexibly within the state space. After taking an incorrect action, the agent can return to a previous state and retry. SheetCopilot \citep{li2024sheetcopilot} utilizes a state machine mechanism to guide the model in re-planning actions by providing error feedback and spreadsheet state feedback, while MobA \citep{zhu2024moba} uses a tree-like task structure to record the complete path, ensuring an efficient backtracking process.
}

\item{
    \textbf{Experience Retrieval.} 
OS Agents can efficiently plan and execute by retrieving experiences similar to the current task from long-term memory, which helps to reduce redundant operations \citep{zheng2023synapse,deng2024multi}. For instance, AWM \citep{wang2024agent} extracts similar task workflows from past tasks and reuses them in new tasks, minimizing the need for repetitive learning.  Additionally, PeriGuru \citep{fu2024periguru} uses the K-Nearest Neighbors algorithm to retrieve similar task cases from a task database and combines them with Historical Actions to enhance decision-making through prompts. 
}
\end{itemize}

\subsubsection{Action}
\label{sec:action}
The action space defines the interfaces through which (M)LLM-based Agents engage with operating systems, spanning across platforms such as computers, mobile devices, and web browsers. We systematically categorized the action space of OS Agents into input operations, navigation operations, and extended operations.

\textbf{Input Operations.} Input operations encompass interactions via mouse/touch and keyboard, forming the foundation for OS Agents to interact with digital interfaces.

Mouse and touch operations encompass three primary types: (1) \textit{click/tap} actions that are universally implemented across different platforms and serve as the most basic form of interaction \cite{sun2022meta,deng2024mind2web,zheng2023synapse}, (2) \textit{long press/hold} actions that are particularly crucial for mobile interfaces and context menu activation \cite{zhang2023appagent,rawles2024androidworld,fu2024periguru}, and (3) \textit{drag/move} operations that enable precise control and manipulation of interface elements \cite{gao2023assistgui,niu2024screenagent,cho2024caap}.

Keyboard operations comprise two main categories: (1) \textit{basic text input} capabilities that allow agents to enter alphanumeric characters and symbols \cite{sun2022meta,deng2024mind2web,zhang2023you}, and (2) \textit{special key} operations (e.g., shortcuts, function keys) \cite{sun2022meta,gao2023assistgui,bonatti2024windows} that enable agents to efficiently navigate and manipulate target applications through keyboard commands.

\textbf{Navigation Operations.} Navigation operations enable OS Agents to traverse targeted platforms and acquire sufficient information for subsequent actions. Navigation operations encompass both basic navigation and web-specific navigation features.

Basic navigation includes: (1) \textit{scroll} operations that enable agents to explore content beyond the current viewport, particularly crucial for processing long documents or infinite-scroll interfaces \cite{yan2023gpt,lee2023explore,gao2023assistgui}, (2) \textit{back/forward navigation} that allows agents to traverse through navigation history and return to previously visited states \cite{sun2022meta,zhang2023you,zhang2023appagent}, and (3) \textit{home function} that provides quick access to the initial or default state of applications, ensuring reliable reset points during task execution \cite{zhang2023you,zhang2023appagent,wang2024mobile}.

Web navigation extends these capabilities with (1) \textit{tab management} that enables agents to handle multiple concurrent sessions and switches between different web contexts \cite{koh2024tree,he2024webvoyager,song2024beyond}, and (2) \textit{URL navigation} features that allow direct access to specific web resources and facilitate efficient web traversal \cite{he2024webvoyager,deng2024mind2web,ma2023laser}.

\textbf{Extended Operations.} Extended Operations provide additional capabilities beyond standard interface interactions, enabling more flexible and powerful agent behaviors. These operations primarily include (1) \textit{code execution} capabilities that allow agents to dynamically extend their action space beyond predefined operations, enabling flexible and customizable control through direct script execution and command interpretation \cite{wu2024copilot,mei2024aios,tan2024cradle}, and (2) \textit{API integration} features that expand agents' capabilities by accessing external tools and information resources, facilitating interactions with third-party services and specialized functionalities \cite{wu2024copilot,mei2024aios,tan2024cradle,li2024sheetcopilot}. These operations fundamentally enhance the adaptability and functionality of OS Agents, allowing them to handle more complex and diverse tasks that may not be achievable through conventional interface-based interactions alone.

\section{Evaluation of OS Agents}
\label{sec:evaluation}
Evaluation plays a crucial role in developing OS Agents, as it helps assess their performance and effectiveness in various scenarios.
The current literature features a multitude of evaluation techniques, which vary significantly according to the specific environment and application. For a clear display and summary of the evaluation framework, we will delve into a comprehensive overview of a generic evaluation framework for OS Agents, structured around evaluation protocols and benchmarking. At the same time, we have provided the recent benchmarks for OS Agents in Table \ref{tab:table_benchmark}.

\begin{table}
\centering
\caption{Recent benchmarks for OS Agents. We divided the Benchmarks into three sections based on the Platform (as mentioned in \S \hyperref[sec:evaluation_platforms]{4.2.1}) and sorted them by release date. The following is an explanation of the abbreviations. BS: Benchmark Settings, M/P: Mobile, PC: Desktop, IT: Interactive, ST: Static, OET: Operation Environment Types, RW: Real-World, SM:Simulated, GG: GUI Grounding, IF: Information Processing, AT:Agentic, CG: Code Generation.}
\label{tab:table_benchmark}
\begin{tblr}{
  colspec = {lcccccc},
  cells = {c},         %
  column{1} = {l},     %
  hline{1,31} = {-}{0.08em},
  hline{2,12,17} = {-}{0.05em},
}
\textbf{Benchmark}        &\textbf{Platform}           & \textbf{BS} & \textbf{OET }  & \textbf{Task}              & \textbf{Date}    \\
AndroidControl \citep{lieffects}  & M/P                 & ST        & - & AT                  & 06/2024 \\
AndroidWorld \citep{rawles2024androidworld}     & M/P      & IT         & RW & AT                  & 05/2024 \\
Android-50 \citep{bishop2024latent}& M/P      & IT         & RW & AT                  & 05/2024 \\
B-MoCA \citep{lee2024benchmarking}          & M/P      & IT         & RW & AT                  & 04/2024 \\
LlamaTouch \citep{zhang2024llamatouch}      & M/P                 & IT         & RW & AT                  & 04/2024 \\
AndroidArena \citep{venkatesh2022ugif}    & M/P                 & IT         & RW & AT                  & 02/2024 \\
AITW \citep{rawles2024androidinthewild}            & M/P      & ST       & - & AT                  & 07/2023 \\
UGIF-DataSet \citep{venkatesh2022ugif}    & M/P                 & ST        & - & AT                  & 11/2022 \\
MoTIF \citep{burns2022dataset}           & M/P      & ST        & - & AT                  & 02/2022 \\
PIXELHELP \citep{li2020mapping}        & M/P      & IT         & RW & GG                   & 05/2020 \\

WindowsAgentArena \citep{bonatti2024windows} & PC    & IT         & RW & AT                  & 09/2024 \\
OfficeBench \citep{wang2024officebench}    & PC    & IT         & RW & AT                  & 07/2024 \\
OSWorld \citep{xie2024osworld}        & PC    & IT         & RW & AT                 & 04/2024 \\
OmniACT \citep{kapoor2024omniact}        & PC    & ST        & - & CG             & 02/2024 \\
ASSISTGUI \citep{gao2023assistgui}      & PC    & IT         & RW & AT                  & 12/2023 \\

Mind2Web-Live \citep{pan2024webcanvas}  & Web            & IT        & RW & IF, AT & 06/2024 \\
MMInA \citep{zhang2024mmina}          & Web            & IT         & RW & IF, AT & 04/2024 \\
GroundUI \citep{zheng2024agentstudio}       & Web            & ST         & - & GG                  & 03/2024 \\
TurkingBench \citep{xu2024tur}    & Web            & IT         & RW & AT                  & 03/2024 \\
WorkArena \citep{drouin2024workarena}      & Web     & IT         & RW & IF, AT & 03/2024 \\
WebLINX \citep{lu2024weblinx}        & Web            & ST        & - & IF, AT & 02/2024 \\
Visualwebarena \citep{koh2024visualwebarena}  & Web   & IT         & RW & GG, AT  & 01/2024 \\
WebVLN-v1 \citep{chen2024webvln}      & Web            & IT        & RW & IF, AT & 12/2023 \\
WebArena \citep{zhou2023webarena}       & Web            & IT         & RW & AT                  & 07/2023 \\
Mind2Web \citep{deng2024mind2web}       & Web            & ST        & - & IF, AT& 06/2023 \\
WebShop \citep{yao2022webshop}        & Web            & ST        & -  & AT & 07/2022 \\
PhraseNode \citep{pasupat2018mapping}     &Web            & ST        & - & GG   & 08/2018 \\
MiniWoB \citep{shi2017world}        & Web & ST     & - & AT    & 08/2017 \\
FormWoB \citep{shi2017world}        & Web & IT       & SM  & AT    & 08/2017 
\end{tblr}
\end{table}

\subsection{Evaluation Protocol}
\label{sec:evaluation_protocol}
This section is dedicated to outlining the comprehensive evaluation protocols. Central to the assessment of OS Agents are two pivotal concerns: (1) \textit{Evaluation Principles}: how the evaluation process should be conducted, and (2) \textit{Evaluation Metrics}: which aspects need to be assessed. We will now elaborate on the principles and metrics for evaluating OS Agents, focusing on these two issues.

\subsubsection{Evaluation Principle}
The evaluation of OS Agents requires a combination of multiple aspects and techniques to gain a comprehensive insight into their capabilities and limitations. The assessment process can be primarily divided into objective and subjective evaluations. This integration of objective and subjective evaluation methods not only secures the assessment of performance in controlled environments, but also prioritizes the agent's reliability and practical usability in real-world situations.

\textbf{Objective Evaluation.}
Objective evaluation primarily measures the performance of OS Agents based on standardized numerical metrics, which are typically rule-based calculations or hardcoded assessments on standard benchmark datasets. 
This form of evaluation specifically targets the agent's accuracy in perception \citep{wang2024webquest,ying2024mmt}, the quality of its generated content \citep{jin2024shopping,xu2024tur}, the effectiveness of its actions \citep{xu2024androidlab}, and its operational efficiency \citep{lee2024benchmarking,wang2024mobileagentbench}. 
Typically, the computation of specific metrics encompasses exact match \citep{xu2024tur,pan2024webcanvas}, fuzzy match \citep{zhang2024mmina}, and semantic matching for text, elements, and images.
Through precise and efficient numerical analysis, objective evaluation enables quick and standardized measurement of the agent's performance.

\textbf{Subjective Evaluation.}
Besides automated objective assessments, subjective evaluations are also essential. These human-centered subjective evaluations aim to measure how well the output matches human expectations \citep{yan2023gpt,pan2024webcanvas,xu2024androidlab}, typically applied in scenarios that require a high level of comprehension and are difficult to quantify using traditional metrics. Such subjective evaluations are based on different subjective aspects, including relevance, coherence, naturalness, harmlessness, and overall quality. 
Early subjective evaluations were primarily based on direct human assessments \citep{DBLP:conf/nips/ZhengC00WZL0LXZ23}, which, while yielding high-quality results, are expensive and difficult to reproduce. Later, LLMs were introduced as evaluators to substitute for human judgment \citep{liu2023agentbench,vu2024foundationalautoraterstaminglarge}, exploiting their strong instruction-following capabilities. 
Such LLM-as-a-judge evaluation method \citep{gu2024surveyllmasajudge,kim2024prometheusinducingfinegrainedevaluation,kim2024prometheus2opensource} can offer detailed explanations for annotation, providing a finer-grained understanding of the agent's strengths and weaknesses.  Nevertheless, despite the gains in efficiency, there are still limitations regarding its reliability and controllability \citep{pasupat2018mapping,gou2024navigating,dardouri2024visual}.

\subsubsection{Evaluation Metric}
As mentioned in \S \ref{sec:capabilities}, the evaluation process of OS Agents mainly examines their abilities in terms of understanding, planning and action grounding.
During evaluation, the agent, provided with task instructions and the current environment input, is expected to execute a sequence of continuous actions until the task is accomplished. By collecting the agent's observations, action outputs, and other environmental information during the process, specific metrics can be calculated. 
Specifically, the evaluation scope includes both granular step-level evaluations and a more holistic task-level assessment. The former focuses on whether each step in the process aligns with the predefined path, while the latter is concerned with whether the agent achieves the goal in the end.

\textbf{Step-level Evaluation.}
Step-level evaluation centers on a detailed, step-by-step analysis of the planning trajectory, offering a fine-grained evaluation of the actions taken by the agent at each step.
In step-level evaluation, the agent's output in response to instruction of each step is directly assessed, with a focus on the accuracy of action grounding and the matching of potential object elements (which refers to the target of the action).
For action grounding, the predicted action at each step is typically compared directly with the reference action to obtain operation metrics, such as \textit{operation accuracy and F1} \citep{xu2024androidlab,jin2024shopping}. For element matching of actions, different approaches are used depending on the type of action and elements, for example, comparing based on element ID or the element position, leading to \textit{element accuracy and F1} \citep{pasupat2018mapping}. In the case of specific tasks, such as those involving visual grounding in question-answering, there are dedicated metrics like \textit{BLEU} \citep{jin2024shopping}, \textit{ROUGE} \citep{xu2024tur}, and \textit{BERTScore} \cite{DBLP:journals/corr/abs-2406-10300}. By aggregating all the relevant metrics for a single step, it is possible to assess the step's success, thereby obtaining the \textit{step success rate (step SR)} \citep{pan2024webcanvas}. 
Despite providing fine-grained comprehension, such step-level evaluation has limitations in assessing the performance of long, continuous action sequences \citep{koh2024visualwebarena,pasupat2018mapping,xie2024osworld}, and a given task may have various valid paths. To boost the robustness \citep{zhang2024pptc} of the evaluation, it is usually necessary to integrate the final task outcome into the assessment.

\textbf{Task-level Evaluation.}
 Task-level evaluation centers on the final output and evaluates whether the agent reaches the desired final state. The two main criteria are task completion and resource utilization. The former assesses whether the agent has successfully fulfilled the assigned tasks as per the instructions, while the latter examines the agent's overall efficiency during task completion.

\begin{itemize}[leftmargin=*]
    \item {
    \textbf{Task Completion Metrics.} Task Completion Metrics measure the effectiveness of OS Agents in successfully accomplishing assigned tasks. These metrics cover several key aspects. \textit{Overall Success Rate (SR)} \citep{koh2024visualwebarena,zhang2023you, drouin2024workarena, shi2017world} provides a straightforward measure of the proportion of tasks that are fully completed. \textit{Accuracy} \citep{ying2024mmt,wang2024webquest,zhang2024pptc} assesses the precision of the agent's responses or actions, ensuring outputs closely match with the expected outcomes. Additionally, \textit{Reward function} \citep{koh2024visualwebarena,yao2022webshop,zhang2023mobile,kapoor2024omniact} is another critical metric, which assigns numerical values to guide agents toward specific objectives in reinforcement learning.
    }

    \item {
    \textbf{Efficiency Metrics.} Efficiency Metrics evaluate how efficiently the agent completes assigned tasks, considering factors such as step cost, hardware expenses, and time expenditure. Specifically, \textit{Step Ratio} \citep{chen2024spa,lee2024benchmarking,wang2024mobileagentbench} compares the number of steps taken by the agent to the optimal one (often defined by human performance). A lower step ratio indicates a more efficient and optimized task execution, while higher ratios highlight redundant or unnecessary actions.
    \textit{API Cost} \citep{guo2023pptc,zhang2024pptc,deng2024mobile} evaluates the financial costs associated with API calls, which is particularly relevant for agents that use external language models or cloud services. 
    Furthermore,  \textit{Execution Time} \citep{xu2024crab} measures the time required for the agent to complete a task, and \textit{Peak Memory Allocation} \citep{zhang2024mmina} shows the maximum GPU memory usage during computation.
    These efficiency metrics are critical for evaluating the real-time performance of agents, especially in resource-constrained environments.   
    }
    
\end{itemize}

\subsection{Evaluation Benchmark}
\label{sec:evaluation_benchmark}
To comprehensively evaluate the performance and capabilities of OS Agents, researchers have developed a variety of benchmarks. These benchmarks construct various environments, based on different platforms and settings, and cover a wide range of tasks. 
This subsection offers a detailed overview of these benchmarks, organized by evaluation platforms, benchmark settings, and tasks.
\subsubsection{Evaluation Platform}
\label{sec:evaluation_platforms}
The platform acts as an integrated evaluation environment, specifically encompassing the virtual settings in which benchmarks are performed. 
Different platforms present unique challenges and evaluation focuses. Some benchmarks also incorporate multiple platforms at the same time, which places greater demands on the agent's cross-platform transferability. Existing real-world platforms can primarily be categorized into three types: Mobile, Desktop, and Web. Each platform has its unique characteristics and evaluation focuses, which we will elaborate on as follows.

\textbf{Mobile.} Mobile platforms such as Android \citep{li2024effects,lee2024benchmarking,bishop2024latent,venkatesh2022ugif} or iOS \citep{yan2023gpt} present unique challenges for OS Agents. While mobile GUI elements are simpler due to smaller screens, they require more complex actions, such as precise gestures for navigating widgets or zooming. The open nature of Android provides a wider action space, encompassing standard GUI interactions and function-calling APIs, such as sending text messages, which imposes higher demands on the agents' planning and action grounding capabilities.

\textbf{Desktop.} Desktop platform is more complex due to the diversity of operating systems and applications. Efficient desktop benchmarks \citep{xie2024osworld,wang2024officebench,bonatti2024windows} need to handle the wide variety and complexity of real-world computing environments, which span different operating systems, interfaces, and applications. As a result, the scope of manageable tasks and the scalability of testing agents are often constrained.

\textbf{Web.} Web platforms are essential interfaces to access online resources. Webpages \citep{koh2024visualwebarena,lu2024weblinx,drouin2024workarena,yao2022webshop,shi2017world} are open and built with HTML, CSS, and JavaScript, making them easy to inspect and modify in real-time. Since agents interact with the web interface in the same way humans do, it’s possible to crowdsource human demonstrations of web tasks from anyone with access to a web browser, keyboard, and mouse, at a low cost. This accessibility has also attracted significant attention from researchers in the field.

\subsubsection{Benchmark Setting}
Apart from the categorization of platforms, the environmental spaces for OS Agents to percept and take actions vary across different evaluation benchmarks. We have organized the existing benchmark environments, primarily dividing them into \textbf{static} and \textbf{interactive} categories, with the interactive environments further split into \textbf{simulated} and \textbf{real-world} settings.

        \textbf{Static.} Static Environments, which are prevalent in early studies, are often created by caching website copies or static data, thereby establishing an offline context for evaluation. The process of setting up a static environment is quite simple, as it merely involves caching the content from real websites. Evaluations generally rely on the cached static content for tasks such as visual grounding, and only one-step action are supported.
        MiniWoB \citep{shi2017world} is built on simple HTML/CSS/JavaScript pages and employs predefined simulation tasks. Mind2Web \citep{deng2024mind2web} captures comprehensive snapshots of each website along with complete interaction traces, enabling seamless offline replay. 
        Owing to the lack of dynamic interaction and environmental feedback, such static evaluations tend to be less authentic and versatile, making them inadequate for a comprehensive assessment.

        \textbf{Interactive.} Interactive Environments provide a more authentic scenario, characterized by their dynamism and interactivity. In contrast to static environments, OS Agents can execute a sequence of actions, receive feedback from the environment, and make corresponding adjustments. Interactive evaluation settings facilitate the evaluation of an agent’s skills in more sophisticated settings. These interactive environments can be subdivided into simulated and real-world types.
        (1) For the \textit{simulated environment}, FormWoB \citep{shi2017world} created a virtual website to avoid the reproducibility issues caused by the dynamic nature of real-world environments, while \citet{rawles2024androidinthewild} developed virtual apps to assess the capabilities of OS Agents. However, these simulated environments are often overly simplistic by excluding unexpected conditions, thus failing to capture the complexity of real-world scenarios. 
        (2) For the \textit{real-world environment}, which is truly authentic and encompasses real websites and apps, one must consider the continuously updating nature of the environment, uncontrollable user behaviors, and diverse device setups. This scenario underscores the requirement for agents to exhibit strong generalization across real-world conditions. 
        OSWorld \citep{xie2024osworld}, for example, constructed virtual machines running Windows, Linux, and MacOS to systematically evaluate the performance of OS Agents across different operating systems. Similarly, AndroidWorld \citep{rawles2024androidworld}, conducted tests on real apps using Android emulators, highlighting the importance of evaluating agents under diverse and realistic conditions.

\subsubsection{Task}

To comprehensively assess the capabilities of OS Agents, a spectrum of specialized tasks has been integrated into the established benchmarks. These tasks span from system-level tasks such as installing and uninstalling applications to daily application such as sending emails and shopping online. These tasks are intended to measure how closely current agents can mimic human performance.

\textbf{Task Categorization.} In evaluating OS Agents, task categorization is critical for understanding their capabilities and limitations at a fine-grained level. Based on the capabilities required by the evaluation process, current benchmark tasks can primarily be categorized into three types: \textbf{GUI Grounding}, \textbf{Information Processing} and \textbf{Agentic Tasks}, details of which are described as follows.
\begin{itemize}[leftmargin=*]

    \item {
    \textbf{GUI Grounding.} GUI grounding tasks aim to evaluate agent's abilities to transform instructions to various actionable elements. Grounding is fundamental for interacting with operation systems that OS Agents must possess. Early works, such as PIXELHELP \citep{li2020mapping}, provide a benchmark that pairs English instructions with actions performed by users on a mobile emulator.
    }

    \item {

    \textbf{Information Processing.} In the context of interactive agents, the ability to effectively handle information is a critical component for addressing complex tasks. This encompasses not only retrieving relevant data from various sources but also summarizing and distilling information to meet specific user needs. Such capabilities are particularly essential in dynamic and diverse environments, where agents must process large volumes of information, and deliver accurate results. To explore these competencies, Information Processing Tasks can be further categorized into two main types:
    (1) \textit{Information Retrieval Tasks} \citep{pan2024webcanvas,zhang2024mmina,drouin2024workarena} examine agent's ability to process complex and dynamic information by understanding instructions and GUI interfaces, extracting the desired information or data. Browsers (either web-based or local applications) are ideal platforms for information retrieval tasks due to their vast repositories of information. Additionally, applications with integrated data services also serve as retrieval platforms. For instance, AndroidWorld \citep{rawles2024androidworld} requires OS Agents to retrieve scheduled events from Simple Calendar Pro.
    (2) \textit{Information Summarizing Tasks} are designed to summarize specified information from a GUI interface, testing agent's ability to comprehend and process information. For example, certain tasks in WebLinx \citep{lu2024weblinx} focus on summarizing web-based news articles or user reviews.
    }
    \item {
    \textbf{Agentic Tasks.} Agentic tasks are designed to evaluate an agent's core abilities (as mentioned in \S \ref{sec:capabilities}) and represent a key focus in current research. In these tasks, OS Agents are provided with an instruction or goal and tasked with identifying the required steps, planning actions, and executing them until the target state is reached, without relying on any explicit navigation guidance. For instance, WebLINX \citep{lu2024weblinx} offers both low-level and high-level instructions, challenging agents to complete single-step or multi-step tasks, thereby testing their planning capabilities. Similarly, MMInA \citep{zhang2024mmina} emphasizes multi-hop tasks, requiring agents to navigate across multiple websites to fulfill the given instruction.
    }
\end{itemize}

\section{Challenge \& Future}
\label{sec:challenges_and_future}
\subsection{Safety \& Privacy}
\label{sec:safety_and_privacy}
A recent report \citep{theblock2024human} highlighted a notable case where a human player successfully outwitted the Freysa AI agent in a \$47,000 crypto challenge, underscoring vulnerabilities even in advanced AI systems and emphasizing the need to address these security risks. This incident aligns with broader concerns as (M)LLMs are increasingly integrated into diverse domains, such as healthcare, education, and autonomous systems, where security has become a critical issue. This growing adoption has led to numerous studies \citep{deng2024aiagentsthreatsurvey, gan2024navigatingriskssurveysecurity, Yao_2024, shayegani2023surveyvulnerabilitieslargelanguage, cui2024risktaxonomymitigationassessment, wang2024boosting, neel2024privacyissueslargelanguage} investigating the security risks associated with LLMs and their applications. In particular, some research has delved into the challenges faced by OS Agents regarding security risks. The following subsections discuss existing research on the security aspects of OS Agents. \S \ref{sec:safety_and_privacy_attack} analyzes various attack strategies targeting OS Agents, \S \ref{sec:safety_and_privacy_defense} explores existing defense mechanisms and limitations, and \S \ref{sec:safety_and_privacy_benchmark} reviews existing security benchmarks designed to assess the robustness and reliability of OS Agents.

\subsubsection{Attack}
\label{sec:safety_and_privacy_attack}
Several researchers have investigated adversarial attacks targeting OS Agents. \citet{wu2024wipinewwebthreat} identified a novel threat called Web Indirect Prompt Injection (WIPI), in which adversaries indirectly control LLM-based Web Agents by embedding natural language instructions into web pages. Recent findings \citep{wu2024adversarialattacksmultimodalagents} further uncovered security risks for MLLMs, illustrating how adversaries can generate adversarial images that cause the captioner to produce adversarial captions, ultimately leading the agents to deviate from the user's intended goals. Similar vulnerabilities have been identified in other studies. \citet{ma2024cautionenvironmentmultimodalagents} introduced an attack method called environmental injection, highlighting that advanced MLLMs are vulnerable to environmental distractions, which can cause agents to perform unfaithful behaviors. Expanding on the concept, \citet{liao2024eiaenvironmentalinjectionattack} executed an environmental injection attack by embedding invisible malicious instructions within web pages, prompting the agents to assist adversaries in stealing users' personal information. \citet{xu2024advwebcontrollableblackboxattacks} further advanced this approach by leveraging malicious instructions generated by an adversarial prompter model, trained on both successful and failed attack data, to mislead MLLM-based Web Agents into executing targeted adversarial actions.

Other studies have explored security issues in specific environments. \citet{zhang2024attackingvisionlanguagecomputeragents} explored adversarial pop-up window attacks on MLLM-based Web Agents, demonstrating how this method interferes with the decision-making process of the agents. \citet{kumar2024refusaltrainedllmseasilyjailbroken} investigated the security of refusal-trained LLMs when deployed as browser agents. Their study found that these models' ability to reject harmful instructions in conversational settings does not effectively transfer to browser-based environments. Moreover, existing attack methods can successfully bypass their security measures, enabling jailbreaking. \citet{yang2024securitymatrixmultimodalagents} proposed a security threat matrix for agents running on mobile devices, systematically examining the security issues of MLLM-based Mobile Agents and identifying four realistic attack paths and eight attack methods.

\subsubsection{Defense}
\label{sec:safety_and_privacy_defense}
Although several security frameworks have been developed for LLM-based Agents \citep{ruan2024identifyingriskslmagents, hua2024trustagentsafetrustworthyllmbased, fang2024inferactinferringsafeactions, xiang2024guardagentsafeguardllmagents, shamsujjoha2024designingmultilayeredruntimeguardrails}, studies on defenses specific to OS Agents \citep{pedro2023promptinjectionssqlinjection} remain limited. Bridging this gap requires the development of robust defense mechanisms tailored to the vulnerabilities of OS Agents, such as injection attacks, backdoor exploits, and other potential threats. Future research could prioritize these areas, focusing on developing comprehensive and scalable security solutions for OS Agents.

\subsubsection{Benchmark}
\label{sec:safety_and_privacy_benchmark}
Several security benchmarks \citep{levy2024stwebagentbenchbenchmarkevaluatingsafety, lee2024mobilesafetybenchevaluatingsafetyautonomous} have been introduced to evaluate the robustness of OS Agents in various scenarios. The online benchmark ST-WebAgentBench \citep{levy2024stwebagentbenchbenchmarkevaluatingsafety} has been developed to systematically assess the safety and trustworthiness of web agents within enterprise environments. It focuses on six key dimensions of reliability, offering a comprehensive framework for evaluating agent behavior in high-risk contexts. Similarly, a benchmarking platform named MobileSafetyBench \citep{lee2024mobilesafetybenchevaluatingsafetyautonomous} has been developed to assess the security of LLM-based Mobile Agents, focusing on evaluating their performance in handling safety-critical tasks within Android environments, including interactions with messaging and banking applications.

\subsection{Personalization \& Self-Evolution}
\label{sec:personalization_and_self_evolution}
Much like Jarvis as Iron Man's personal assistant in the movies, developing personalized OS Agents has been a long-standing goal in AI research. A personal assistant is expected to continuously adapt and provide enhanced experiences based on individual user preferences. OpenAI's memory feature\footnote{\url{https://openai.com/index/memory-and-new-controls-for-chatgpt/}} has made strides in this direction, but many (M)LLMs today still perform insufficient in providing personalized experience to users and self-evolving over user interactions. 

Early works \citep{wang2023voyager,zhu2023ghost} allowed LLM-based Agents to interact with environments of games, summarizing experiences into text, thus accumulating memory and facilitating self-evolution~\citep{zhou2024symboliclearning}. For example, \citet{wang2023voyager} demonstrated the potential for agents to adapt and evolve through experience. Later, researchers applied these principles to the OS Agent domain \citep{zhang2023appagent,li2024appagent,wu2024copilot}. These efforts validated the feasibility of memory mechanisms in OS Agents. Although due to the limited resources available in academia and the difficulty of accessing real user data, much of the current research focuses on improving performance for specific tasks rather than personalization. The memory mechanism still shows potential for OS Agents to accumulate user data over time, thus improving user experience and performance.

Moreover, expanding the modalities of memory from text to other forms, such as images, voice, presents significant challenges. Managing and retrieving this memory effectively also remains an open issue. We believe that in the future, overcoming these challenges will enable OS Agents to provide more personalized, dynamic, and context-aware assistance, with more sophisticated self-evolution mechanisms that continually adapt to the user's needs and prefernces.

\section{Related Work}
(Multimodal) Large Language Models \citep{wake2024yi,li2024ariaopenmultimodalnative,zheng2024minigpt5interleavedvisionandlanguagegeneration,bai2023qwenvlversatilevisionlanguagemodel,dai2022modelmultiplemodalitiessparsely} have emerged as transformative tools in artificial intelligence, driving significant advancements across various domains. \citet{zhao2023survey} summarize a foundational overview of LLMs. \citet{yin2024survey,zhang2024mm} comprehensively reviews the progress of Multimodal LLMs. In addtion, \citet{long2024llms} explores the use of synthetic data for training. \citet{zhang2023instruction} presents the current state of research on the field of instruction tuning for LLMs.

With the flourishing development of (M)LLM-based Agents, numerous comprehensive surveys have emerged, offering detailed insights into various aspects of these systems. \citet{wang2024survey,cheng2024exploring,gan2024navigating} provides an overview of general LLM-based Agents. For the agent frameworks, \citet{zhou2023agents,zhang2024survey,li2024survey} explore methods to enhance agents' capabilities of planning, memory and multi-agents interaction. \citet{qiao2022reasoning} presents comprehensive comparisons for LLM’s reasoning abilities. \citet{hou2023large,hu2024survey,li2024personal} summarizes studies in different application fields including software engineering, game and personal assistance. Some concurrent works \citep{li2024personalllmagentsinsights,wu2024foundationsrecenttrendsmultimodal,wang2024guiagentsfoundationmodels,gao2024generalistvirtualagentssurvey,zhang2024large} touch on concepts that share some features with OS Agents, such as personalized agents, GUI Agents and generalist virtual agents. This work aims to provide an integrated view on the construction and evaluation of OS Agents, that leverage environments and interfaces provided by operating systems, while identifying open challenges and future directions in this domain for forthcoming studies.

\section{Conclusion}
The development of (multimodal) large language models has created new opportunities for OS Agents, moving the idea of advanced AI assistants closer to being realized. In this survey, we have aimed to outline the fundamentals underlying OS Agents, including their key components and capabilities. We have also reviewed various approaches to their construction, with particular attention to domain-specific foundation models and agent frameworks. Through the evaluation protocols and benchmarks discussed, we have explored methods for assessing the performance of OS Agents across a variety of tasks. Looking ahead, we identify critical challenges, such as safety and privacy, personalization and self-evolution, as areas that require continued research and attention. This summary of the current state of the field, along with potential directions for future work, is intended to contribute to the ongoing development of OS Agents and support their relevance and utility in both academic and industrial settings.

\bibliography{custom}
\bibliographystyle{unsrtnat}

\end{document}